# On the Complexity of Strong and Epistemic Credal Networks

Denis D. Mauá      Cassio P. de Campos      Alessio Benavoli      Alessandro Antonucci

Istituto Dalle Molle di Studi sull'Intelligenza Artificiale (IDSIA),
Manno-Lugano, Switzerland,
{denis,cassio,alessio,alessandro}@idsia.ch

## Abstract

Credal networks are graph-based statistical models whose parameters take values in a set, instead of being sharply specified as in traditional statistical models (e.g., Bayesian networks). The computational complexity of inferences on such models depends on the irrelevance/independence concept adopted. In this paper, we study inferential complexity under the concepts of epistemic irrelevance and strong independence. We show that inferences under strong independence are NP-hard even in trees with ternary variables. We prove that under epistemic irrelevance the polynomial time complexity of inferences in credal trees is not likely to extend to more general models (e.g. singly connected networks). These results clearly distinguish networks that admit efficient inferences and those where inferences are most likely hard, and settle several open questions regarding computational complexity.

## 1 INTRODUCTION

Bayesian networks are multivariate statistical models where irrelevance assessments between sets of variables are concisely described by means of an *acyclic directed graph* whose nodes are identified with variables (Pearl, 1988). The graph encodes a set of *Markov conditions*: non-descendant non-parent variables are irrelevant to a variable given its parents. The complete specification of a Bayesian network requires the specification of a conditional probability distribution for every variable and every configuration of its parents. This is no easy task. Whether these parameters (i.e., the conditional distributions) are estimated from data or elicited from experts, they inevitably contain imprecisions and arbitrariness (Kwisthout and van der Gaag, 2008). Despite this fact, Bayesian networks have been successfully applied to a wide range of applications (Koller and Friedman, 2009).

An arguably more principled approach to coping with imprecision in parameters is by means of closed and convex sets of probability mass functions called *credal sets* (Levi, 1980). Unlike the representation of knowledge by "precise" probability mass functions, credal sets allow for the distinction between randomness and ignorance (Walley, 1991). Bayesian networks whose parameters are specified by conditional credal sets are known as *credal networks* (Cozman, 2000). Credal networks have been successfully applied to knowledge-based expert systems, where it has been argued that allowing parameters to be imprecisely specified facilitates elicitation from experts (Antonucci et al., 2007, 2009; Piatti et al., 2010).

A Bayesian network provides a concise representation of the (single) multivariate probability mass function that is consistent with the network parameters and factorizes over the graph. Analogously, a credal network provides a concise representation of the credal set of multivariate mass functions that are consistent with the local credal sets and satisfy (at least) the irrelevances encoded in the graph. The precise characterization of this joint credal set depends however on the concept of irrelevance adopted.

The two most commonly used irrelevance concepts in the literature are *strong independence* and *epistemic irrelevance*. Two variables $X$ and $Y$ are *strongly independent* if the joint credal set of $X, Y$ can be regarded as originating from a number of precise probability mass functions in each of which the two variables are stochastically independent. Strong independence is thus closely related to the sensitivity analysis interpretation of credal sets, which regards an imprecisely specified model as arising out of partial ignorance of an ideal precisely specified one (Kwisthout and van der Gaag, 2008; Antonucci and Piatti, 2009; Zaffalon and Miranda, 2009). A variable $X$ is *epistemically irrel-*

*evant* to a variable $Y$ if the marginal credal set of $Y$ according to our model is the same whether we observe the value of $X$ or not (Walley, 1991).

Typically, credal networks are used to derive tight bounds on the expectation of some variable conditional on the value of some other variables. The complexity of such an inferential task varies greatly according to the topology of the underlying graph, the cardinality of the variable domains, and the irrelevance concept adopted. For instance, the 2U algorithm of Fagiuoli and Zaffalon (1998) can solve the problem in polynomial time if the underlying graph is singly connected, variables are binary and strong independence is assumed. When instead epistemic irrelevance is adopted, no analogous polynomial-time algorithm for the task is known. On the other hand, de Cooman et al. (2010) developed a polynomial-time algorithm for inferences in credal trees under epistemic irrelevance that work for arbitrarily large variable domains. No such algorithm is known under strong independence. Recently, Mauá et al. (2012) showed the existence of a fully polynomial-time approximation scheme for credal networks of bounded treewidth and bounded variable cardinality under strong independence. It is still unknown whether an analogous result can be obtained under epistemic irrelevance.

In this paper, we show that epistemic irrelevance and strong independence induce the same upper and lower predictive probability in HMM-like (hidden Markov model-like) credal networks (i.e., when we query the value of the "last" state node given some evidence), and that they induce the same marginal expectation bounds in any network where non-root nodes are precise and root nodes are vacuous. The former implies that we can use the algorithm of de Cooman et al. (2010) for epistemic credal trees to compute tight bounds on the predictive probability in HMM-like credal networks also under strong independence. The latter implies that computing tight posterior probability bounds in singly connected credal networks over ternary variables under epistemic irrelevance is NP-hard, as we show this to be the case under strong independence (de Campos and Cozman (2005) have previously shown the NP-hardness of inference in singly connected strong networks when variable cardinalities are unbounded). Table 1 summarizes both the previously known complexity results and the contributions of this work, which appear in boldface.

## 2 CREDAL NETWORKS

Let $X_1, \ldots, X_n$ be variables taking values $x_1, \ldots, x_n$ in finite sets $\mathcal{X}_1, \ldots, \mathcal{X}_n$, respectively. We write $X := (X_1, \ldots, X_n)$ to denote the $n$-dimensional variable taking values $x$ in $\mathcal{X} := \mathcal{X}_1 \times \cdots \times \mathcal{X}_n$. Given $X$ and $I \subseteq \{1, \ldots, n\} := N$, the notation $X_I$ denotes the vector obtained from $X$ by discarding the coordinates not in $I$. We also write $\mathcal{X}_I$ to denote the Cartesian product (in the proper order) of the sets $\mathcal{X}_i$ with $i \in I$, and $x_I$ to denote an element of $\mathcal{X}_I$. Note that $X_N = X$ and $\mathcal{X}_N = \mathcal{X}$.

Let $\mathcal{Z}$ be some finite set (multidimensional or not), and $Z$ a variable taking values in $\mathcal{Z}$. A *probability mass function (pmf)* $p(Z)$ is a non-negative real-valued function on $\mathcal{Z}$ such that $\sum_{z \in \mathcal{Z}} p(z) = 1$. Given a pmf $p$ on $\mathcal{Z}$, we define the *expectation* operator as the functional $\mathbb{E}_p[f] := \sum_{z \in \mathcal{Z}} f(z) p(z)$ that maps every real-valued function $f$ on $\mathcal{Z}$ to a real number. Given a pmf $p(X)$ on $\mathcal{X}$, and disjoint subsets $I$, $J$ and $K$, we say that variables $X_K$ are *stochastically irrelevant* to $X_I$ given $X_J$ if $p(x_I|x_{J \cup K}) = p(x_I|x_J)$ for all $x$, where the conditional pmfs are obtained by application of Bayes' rule on $p(X)$. Variables $X_I$ and $X_K$ are independent conditional on $X_J$ if, given $X_J$, $X_I$ and $X_K$ are irrelevant to each other.

A *credal set* $C(Z)$ is a closed and convex set of pmfs $p(Z)$ on $\mathcal{Z}$ (Levi, 1980). The *extrema* of a credal set are the points that cannot be written as convex combinations of other points in the set. The extrema of $C(Z)$ are denoted by $\text{ext}\, C(Z)$. We assume that every credal set has finitely many extrema, which are used to represent it. Thus, the credal sets we consider are geometrically polytopes. The *vacuous credal set* is the largest credal set on a given domain $\mathcal{Z}$, and is denoted by $V(Z)$. Given a set $M(Z)$ of pmfs $p(Z)$ on $\mathcal{Z}$ we write $\text{co}\, M(Z)$ to denote its convex closure, that is, the credal set obtained by all convex combinations of elements of $M(Z)$. Given a credal set $C(X)$, disjoint subsets $I$ and $J$ of $N$, and an assignment $x_J$ to $X_J$, we define the *conditional credal set* $C(X_I|x_J)$ (induced by $C(X)$) as the set $\text{co}\{\, p(X_I|x_J) : p \in C(X), p(x_J) > 0 \,\}$. It can be shown that the $C(X_I|x_J)$ remains the same if we replace $C(X)$ with its extrema $\text{ext}\, C(X)$.

For disjoint subsets $I$, $J$ and $K$ of $N$, we say that a variable $X_K$ is *strongly irrelevant* to $X_I$ given $X_J$ (and w.r.t. $C(X)$) if $X_K$ is stochastically irrelevant to $X_I$ given $X_J$ in every extreme $p \in \text{ext}\, C(X)$. This implies that $C(X_I|x_{J \cup K}) = C(X_I|x_J)$ for all $x$. Variables $X_I$ and $X_K$ are *strongly independent* given $X_J$ if condi-tional on $X_J$ both $X_I$ is strongly irrelevant to $X_K$ and $X_K$ is strongly irrelevant to $X_I$. Since stochastic irrelevance implies stochastic independence, strong independence is implied by strong irrelevance.

Table 1: Inferential Complexity of Credal Networks

| MODEL | STRONG | EPISTEMIC |
|---|---|---|
| (Predictive) HMM | P | P |
| Tree | **NP-hard** | P |
| Singly connected | **NP-hard** | **NP-hard** |
| Multiply connected | $\text{NP}^{\text{PP}}$-hard | **$\text{NP}^{\text{PP}}$-hard** |

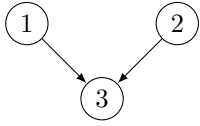

Figure 1: A Simple Polytree

We say that variables $X_K$ are *epistemically irrelevant* to $X_I$ given $X_J$ if $C(X_I|x_{J \cup K}) = C(X_I|x_J)$ for all values of $x$. One can show that strong irrelevance implies epistemic irrelevance (and the converse is not necessarily true) (Cozman, 2000; de Cooman and Troffaes, 2004). Variables $X_I$ and $X_K$ are *epistemically independent* conditional on $X_J$ if, given $X_J$, $X_I$ and $X_K$ are epistemically irrelevant to each other (Walley, 1991, Ch. 9).

Let $G = (N, A)$ be an acyclic directed graph. We denote the *parents* of a node $i$ by $\text{pa}(i)$. The set of *non-descendants* of $i$, written $\text{nd}(i)$, contains the nodes not reachable from $i$ by a directed path. Note that $\text{pa}(i) \subseteq \text{nd}(i)$. We say that $G$ is *singly connected* if there is at most one *undirected* path connecting any two nodes in the graph; it is a *tree* is if additionally each node has at most one parent. If a graph is not singly connected, we say it is *multiply connected*. Singly connected directed graphs are also called *polytrees*.

A (separately specified) credal network $\mathcal{N}$ associates to each node $i$ in $G$ a variable $X_i$ and a collection $Q(X_i|X_{\text{pa}(i)})$ of local credal sets $Q(X_i|x_{\text{pa}(i)})$ indexed by the values of $X_{\text{pa}(i)}$. When every local credal set is a singleton the model specifies a Bayesian network.

**Example 1.** *Consider the credal network $\mathcal{N}$ over Boolean variables $X_1, X_2, X_3$ whose graph is shown in Figure 1. Let $[p(0), p(1)]$ represent a pmf of a Boolean variable. The local credal sets are $Q(X_1) = Q(X_2) = \text{co}\{[0.4, 0.6], [0.5, 0.5]\}$ and $Q(X_3|x_{1,2}) = \{[I(x_1 = x_2), I(x_1 \neq x_2)]\}$, where $I(\cdot)$ is the indicator function.*

The *strong extension* is the credal set $C(X)$ whose extrema $p(X)$ satisfy for all $x$ the condition

$$p(x) = \prod_{i \in N} q(x_i|x_{\text{pa}(i)}), \qquad (1)$$

where $q(X_i|x_{\text{pa}(i)}) \in \text{ext}\, Q(X_i|x_{\text{pa}(i)})$. The strong extension satisfies the Markov condition w.r.t. strong independence: every variable is strongly independent of its non-descendant non-parents given its parents. The *epistemic extension* is the joint credal set $C(X)$ such that

$$C(X_i|x_{\text{nd}(i)}) = Q(X_i|x_{\text{pa}(i)}) \qquad (2)$$

for every variable $X_i$ and value $x_{\text{nd}(i)}$ of $X_{\text{nd}(i)}$. The epistemic extension satisfies the Markov condition w.r.t. epistemic irrelevance: the non-descendant non-parents are irrelevant to a variable given its parents. Equation 2 implies that (and is equivalent to for Boolean variables $X_i$)

$$\min q(x_i|x_{\text{pa}(i)}) \leq p(x_i|x_{\text{nd}(i)}) \leq \max q(x_i|x_{\text{pa}(i)}) \quad (3)$$

for all $x$ and $p \in C(X)$ with $p(x_{\text{nd}(i)}) > 0$, where the optimizations are over $q \in Q(X_i|x_{\text{pa}(i)})$. Note that these inequalities can be turned into linear inequalities by multiplying both sides by $p(x_{\text{nd}(i)})$.

Given a function $f$ of a query variable $X_q$, and an assignment $\tilde{x}_O$ to evidence variables $X_O$, the primary inference with credal networks is the application of the *generalized Bayes rule* (GBR), which asks for a value of $\mu$ that solves the equation

$$\min_{p \in C(X)} \sum_{x \sim \tilde{x}_O} [f(x_q) - \mu] p(x) = 0, \qquad (4)$$

where the sum is performed over the values $x$ of $X$ whose coordinates indexed by $O$ equal $\tilde{x}_O$. Assuming that $\min_{p \in C(X_O)} p(\tilde{x}_O) > 0$, it follows that

$$\mu = \min_{p \in C(X_q|\tilde{x}_O)} \mathbb{E}_p[f], \qquad (5)$$

that is, $\mu$ is the lower expectation of $f$ on the posterior credal set $C(X_q|\tilde{x}_O)$ induced by the (strong or epistemic) extension of the network.

**Example 2.** *Consider the network in Example 1, and let $q = 3$, $f = I(x_3 = 0)$ and $O$ is the empty set. Then applying the GBR is equivalent to finding the lower marginal probability $\mu = \min_{p \in C(X_3)} p(0)$ induced by the network extension. Assuming strong independence (hence strong extension), the inference is solved by $\mu = \min \sum_{x_{1,2}} q_1(x_1) q_2(x_2) q_3(0|x_{1,2}) = 1 + \min\{2q_1(0)q_2(0) - q_1(0) - q_2(0)\} = 1 - 1/2 = 1/2$, where the minimizations are performed over $q_1(X_1) \in \text{ext}\, Q(X_1)$, $q_2(X_2) \in \text{ext}\, Q(X_2)$, and $q_3(X_3|x_{1,2}) = [I(x_1 = x_2), I(x_1 \neq x_2)]$. The epistemic extension is the credal set of joint pmfs $p(X)$ on $\mathcal{X}$ such that*

$$\min_{q \in Q(X_1)} q(x_1) \leq p(x_1|x_2) \leq \max_{q \in Q(X_1)} q(x_1),$$
$$\min_{q \in Q(X_1)} q(x_2) \leq p(x_2|x_1) \leq \max_{q \in Q(X_1)} q(x_2),$$
$$q(x_3|x_{1,2}) \leq p(x_3|x_{\{1,2\}}) \leq q(x_3|x_{1,3}).$$

*The inference under epistemic irrelevance is the value of the solution of the linear program $\mu = \min\{p(0,0,0) + p(1,1,0) : p(X) \in C(X)\} = 5/11 < 1/2$, where $C(X)$ is the epistemic extension.*

# 3 COMPLEXITY RESULTS

In this section we present new results about the computational complexity of GBR inferences in credal networks. We make use of previously unknown equivalences between strong and epistemic extensions to derive both positive and negative complexity results. The section is divided in subsections addressing networks in increasing order of topological complexity.

## 3.1 HIDDEN MARKOV MODELS

An imprecise hidden Markov model (HMM) is a credal network whose nodes can be partitioned into *state* and *manifest* nodes such that the state nodes form a chain (i.e., a sequence of nodes with one node linking to the next and to no other in the sequence), and each manifest node is a leaf with a single state node as parent. As the following example shows, there are GBR inferences in HMMs which depend on the irrelevance concept adopted, even in the case of binary variables.

**Example 3.** *Consider an HMM over four Boolean variables $X_1, X_2, X_3, X_4$, where $X_1$ and $X_2$ are state variables and $X_3$ and $X_4$ are manifest variables. The topology of the corresponding credal network is depicted in Figure 2. The local credal sets are given by $Q(X_1) = Q(X_2|0) = Q(X_4|0) = \{[3/4, 1/4]\}$, $Q(X_2|1) = Q(X_4|1) = \{[1/4, 3/4]\}$, and $Q(X_3|0) = \text{co}\{[1/4, 3/4], [1/2, 1/2]\}$ and $Q(X_3|1) = \text{co}\{[3/4, 1/4], [1/2, 1/2]\}$. Consider the query $f(X_4) = I(x_4 = 0)$, and evidence $\tilde{x}_3 = 0$. Under strong independence, the GBR is to solve for $\mu$ the equation*

$$\min \sum_{x_2} q(0|x_2) g_\mu(x_2) = \sum_{x_2} \min q(0|x_2) g_\mu(x_2) = 0,$$

*where the minimizations are performed over $q(X_3|x_2) \in Q(X_3|x_2)$, $x_2 = 0, 1$, and*

$$g_\mu(x_2) = \sum_{x_{1,4}} q(x_1) q(x_2|x_1) q(x_4|x_1) [I(x_4=0) - \mu],$$

*with $q(X_1) = q(X_2|0) = q(X_4|0) = [3/4, 1/4]$ and $q(X_2|1) = q(X_4|1) = [1/4, 3/4]$. The values of $q(0|x_2)$ depend only on the signs of $g_\mu(x_2)$, $x_2 = 0, 1$. Solving for $\mu$ for each of the four possibilities, and taking the minimum value of $\mu$, we find that $\mu = \min\{p(0|0) : p(X_4|X_3) \in C(X_4|X_3)\} = 4/7$.*

*Under epistemic irrelevance, the GBR is equal to*

$$\min \sum_{x_{1,2,4}} q(x_1) q(x_2|x_1) q(x_4|x_1) q(0|x_{1,2,4}) h_\mu(x_4) =$$

$$(1-\mu) \sum_{x_{1,2}} q(x_1) q(x_2|x_1) q(0|x_1) \min q(0|x_{1,2}, x_4=0)$$

$$-\mu \sum_{x_{1,2}} q(x_1) q(x_2|x_1) q(1|x_1) \max q(0|x_{1,2}, x_4=1) = 0,$$

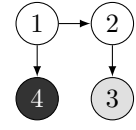

Figure 2: HMM Over Four Variables

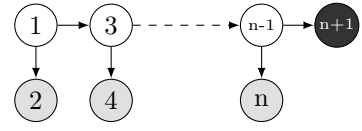

Figure 3: HMM Over $2n+1$ Variables

*where $h_\mu(x_4) = I(x_4 = 0) - \mu$, $q(X_1)$, $q(X_2|X_1)$ and $q(X_4|X_1)$ are defined as before, and $q(X_3|x_{1,2,4}) \in Q(X_3|x_2)$ for every $x_{1,2,4}$. Solving the equation above for $\mu$ we get that $\mu = 13/28$.*

GBR inferences in HMMs are polynomial-time computable under epistemic irrelevance, but no polynomial-time algorithm is known under strong independence except for the case of binary variables (in which case the aforementioned 2U algorithm can be used). The following result shows that there exists a class of GBR inferences in HMMs which are insensitive to the irrelevance concept adopted. This implies that the GBR is polynomial-time computable in such cases also under strong independence.

**Theorem 1.** *Consider an HMM over $n + 1$ variables whose state nodes are identified with odd numbers and manifest nodes are identified with even numbers (see Figure 3). Assume that the query node is $q = n + 1$, and that evidence $\tilde{x}_O$ is set on a subset $O$ of the manifest nodes. Then the posterior lower expectation of any function $f$ on $\mathcal{X}_q$ conditional on $\tilde{x}_O$ is the same whether we assume epistemic irrelevance or strong independence.*

*Proof.* Let $g_\mu(x_{n+1})$ be equal to $f(x_{n+1}) - \mu$. To compute the GBR under strong independence, we need to find a number $\mu$ such that

$$\min \sum_{x \sim \tilde{x}_O} q(x_1) \prod_{i=2}^{n+1} q(x_i | x_{\text{pa}(i)}) g_\mu(x_{n+1}) = 0,$$

where the minimization is performed over $q(x_1) \in Q(X_1)$ and $q(X_i|x_{\text{pa}(i)}) \in Q(X_i|x_{\text{pa}(i)})$ for $i = 2, \ldots, n+1$. The minimization above is equivalent to the constrained program

$$\text{minimize} \quad \sum_{x \sim \tilde{x}_O} q(x_1) \prod_{i=2}^{n+1} q(x_i | x_{1:i-1}) g_\mu(x_{n+1}) \quad (6)$$

$$\text{subject to} \quad q(x_i | x_{1:i-1}) = q(x_i | x_{\text{pa}(i)}),\ i > 2, \quad (7)$$

with variables $q(x_1) \in Q(X_1)$, $q(X_i|x_{\text{pa}(i)}) \in Q(X_i|x_{\text{pa}(i)})$, $i = 2, \ldots, n+1$ and $q(X_i|x_{1:i-1}) \in Q(X_i|x_{\text{pa}(i)})$, $i = 3, \ldots, n+1$. The objective function in (6) can be rewritten as

$$\sum_{x_{1:n} \sim \tilde{x}_O} q(x_1) \prod_{i=2}^{n} q(x_i|x_{1:i-1}) \sum_{x_{n+1}} q(x_{n+1}|x_{1:n}) g_\mu(x_{n+1}).$$

We show the result by induction. Consider a state node $i+1$ and assume that the constrained program (6)–(7) is equivalent to

$$\min_{\text{s.t.}(7)} \sum_{x_{1:i+1} \sim \tilde{x}_O} q(x_1) \prod_{j=2}^{i+1} q(x_j|x_{1:j-1}) h_{i+1}(x_{i+1}),$$

for some function $h_{i+1}$ on $\mathcal{X}_{i+1}$, and let $q^*(X_{i+1}|X_{1:i}) := \{q^*(X_{i+1}|x_{1:i}) : x_{1:i}\}$ be the solutions to the linear optimizations

$$h_i(x_{i-1}) := \min_{q \in Q(X_{i+1}|x_{i-1})} \sum_{x_{i+1}} q(x_{i+1}|x_{1:i}) h_{i+1}(x_{i+1})$$

for different values of $X_{1:i}$. Then $q^*(X_{i+1}|X_{1:i})$ satisfies (7) and minimizes (6) w.r.t. $q(X_{i+1}|X_{1:i})$, thus

$$\min_{\text{s.t.}(7)} \sum_{x_{1:i+1} \sim \tilde{x}_O} q(x_1) \prod_{j=2}^{i+1} q(x_j|x_{1:j-1}) h_{i+1}(x_{i+1}) =$$

$$\min_{\text{s.t.}(7)} \sum_{x_{1:i} \sim \tilde{x}_O} q(x_1) \prod_{j=2}^{i} q(x_j|x_{1:j-1}) h_i(x_{i-1}).$$

Similarly, consider a manifest node $i$ and assume that (6)–(7) is equivalent to

$$\min_{\text{s.t.}(7)} \sum_{x_{1:i} \sim \tilde{x}_O} q(x_1) \prod_{j=2}^{i} q(x_j|x_{1:j-1}) h_i(x_{i-1}). \quad (8)$$

Let $q^*(X_i|X_{1:i-1})$ be the solutions to the linear optimizations

$$h_{i-1}(x_{i-1}) := \min_{q \in Q(X_i|x_{i-1})} \sum_{x_i \sim \tilde{x}_O} q(x_i|x_{1:i-1}) h_i(x_{i-1})$$

for different values of $X_{1:i-1}$. Then, $q^*(X_i|X_{1:i-1})$ satisfies (7) and minimizes (6), therefore (8) equals

$$\min_{\text{s.t.}(7)} \sum_{x_{1:i-1} \sim \tilde{x}_O} q(x_1) \prod_{j=2}^{i-1} q(x_j|x_{1:j-1}) h_{i-1}(x_{i-1}).$$

The basis for $i = n$ follows trivially by setting $h_{n+1}(x_{n+1}) = g_\mu(x_{n+1})$. Thus, the unconstrained minimization in (6) (without the constraints (7)) achieves the same value of the constrained program. Moreover, it can be shown that the unconstrained program is the epistemic extension of the network (Benavoli et al., 2011), so that the result follows. □

**Corollary 1.** *Consider again the HMM of Theorem 1 and the same query setting, and assume that the local credal sets associated to manifest nodes $i$ are singletons $Q(X_i|x_{i-1}) = \{q(x_i|x_{i-1})\}$ such that $q(x_i|x_{i-1}) = 1$ whenever $x_i = x_{i-1}$. Then the posterior lower expectations of any function $f$ given $\tilde{x}_O$ is the same whether we assume epistemic irrelevance or strong independence.*

The proof of Corollary 1 follows directly from Theorem 1. Observe that since $q(x_i|x_{i-1})$ for a manifest node $i$ are 0-1 probabilities, the HMM reduces to a(n imprecise) Markov Chain. Thus, strong and epistemic extensions coincide also in Markov Chains when evidence is before (w.r.t. the topological order) the query node. In the case of evidence after the query, de Cooman et al. (2010) have shown by a counterexample that inferences in Markov Chains are sensitive to the irrelevance concept adopted.

### 3.2 CREDAL TREES

Imprecise HMMs are particular cases of credal trees. De Cooman et al. (2010) showed that GBR inferences can be computed in polynomial time in credal trees under epistemic irrelevance. In this section we show that the same type of inference under strong independence is an NP-hard task.

In the intermediate steps of reductions used to show hardness results we make use of networks whose numerical parameters are specified by *(polynomial-time) computable* numbers, which might not be encodable trivially as rationals. A number $r$ is computable if there exists a machine $M_r$ that, for input $b$, runs in at most time $\text{poly}(b)$ (the notation $\text{poly}(b)$ denotes an arbitrary polynomial function of $b$) and outputs a rational number $t$ such that $|r - t| < 2^{-b}$. Of special relevance are numbers of the form $2^{t_1}/(1 + 2^{t_2})$, with $|t_1|, |t_2|$ being rationals no greater than two, for which we can build a machine that outputs a rational $t$ with the necessary precision in time $\text{poly}(b)$ as follows: compute the Taylor expansions of $2^{t_1}$ and $2^{t_2}$ around zero with sufficiently many terms (depending on the value of $b$), and then compute the fractional expression. The following lemma ensures that any network specified with computable numbers can be approximated arbitrarily well by a network specified with rational numbers.

**Lemma 1.** *Consider a credal network $\mathcal{N}$ over $n$ variables whose numerical parameters $q(x_i|x_{\text{pa}(i)})$ are specified with computable numbers encoded by their respective machines, and let $b$ be the size of the encoding of the network. Given any rational number $\varepsilon \geq 2^{-\text{poly}(b)}$, we can construct in time $\text{poly}(b)$ a credal network $\mathcal{N}'$ over the same variables whose numerical param-*

*eters are all rational numbers, and such that there is a polynomial-time computable bijection $(p, p')$ that associates any extreme $p$ of the strong extension $\mathcal{N}$ with an extreme $p'$ of the strong extension of $\mathcal{N}'$ satisfying $\max_{x_I \in \mathcal{X}_I} |p'(x_I) - p(x_I)| \leq \varepsilon$ for every subset of variables $X_I$.*

*Proof.* Take $\mathcal{N}'$ to be equal to $\mathcal{N}$ except that each computable number $r$, given by its machine $M_r$, used in the specification of $\mathcal{N}$ is replaced by a rational $t$ such that $|t - r| < 2^{-(n+1)(v+1)}\varepsilon$, where $v := \max_{i \in N} |\mathcal{X}_i|$ is the maximum number of values any variable can assume. Because $\varepsilon \geq 2^{-\text{poly}(b)}$, we can use $M_r$ with input $\text{poly}(b) + (n+1)(v+1)$ to obtain $t$ in time $O(\text{poly}(\text{poly}(b) + (n+1)(v+1))) = O(\text{poly}(b))$. Exactly one of the probability values in each pmf in $\mathcal{N}'$ is computed as one minus the sum of the other numbers to ensure that the total mass of the pmf is exactly one; its error is at most $(v-1) \cdot 2^{-(n+1)(v+1)}\varepsilon < 2^{-n(v+1)}\varepsilon$.

Let $q(x_i | x_{\text{pa}(i)})$ and $q'(x_i | x_{\text{pa}(i)})$ denote the parameters of $\mathcal{N}$ and $\mathcal{N}'$, respectively, and consider an assignment $x$ to all variables $X$ in $\mathcal{N}$ (or in $\mathcal{N}'$). By design $|q'(x_i | x_{\text{pa}(i)}) - q(x_i | x_{\text{pa}(i)})| \leq 2^{-n(v+1)}\varepsilon$. It follows from the binomial expansion of the factorization of $p'(x)$ that (there is a term for $p(x)$ in the expansion and $2^n - 1$ terms that an be written as a product of $2^{-n(v+1)}\varepsilon$ by numbers less than or equal to one)

$$p'(x) = \prod_{i \in N} q'(x_i | x_{\text{pa}(i)})$$
$$\leq \prod_{i \in N} [2^{-n(v+1)}\varepsilon + q(x_i | x_{\text{pa}(i)})]$$
$$= \sum_{S \subseteq N} \prod_{i \in S} q(x_i | x_{\text{pa}(i)}) [2^{-n-vn}\varepsilon]^{n-|S|}$$
$$\leq 2^n 2^{-n-vn}\varepsilon + \prod_{i \in N} q(x_i | x_{\text{pa}(i)})$$
$$= p(x) + 2^{-nv}\varepsilon \,.$$

Similarly, we can show that

$$p'(x) \geq \prod_{i=1}^{n} [q(x_i | x_{\text{pa}(i)}) - 2^{-n(v+1)}\varepsilon] \geq p(x) - 2^{-nv}\varepsilon \,.$$

Thus, $\max_x |p'(x) - p(x)| \leq 2^{-nv}\varepsilon$. Now consider a subset of the variables $X_I$ and a value $\tilde{x}_I \in \mathcal{X}_I$. Since $p'(\tilde{x}_I) = \sum_{x \sim \tilde{x}_I} p'(x)$, each term $p'(x)$ in the sum satisfies $p'(x) \leq p(x) + 2^{-nv}\varepsilon$, and there are less than $v^n \leq 2^{vn}$ terms being summed, we have that

$$p'(\tilde{x}_I) \leq \sum_{x \sim \tilde{x}_I} [p(x) + 2^{-vn}\varepsilon] \leq p(\tilde{x}_I) + \varepsilon \,.$$

An analogous argument can be used to show that $p'(\tilde{x}_I) \geq p(\tilde{x}_I) - \varepsilon$. Thus, $\max_{x_I} |p'(x_I) - p(x_I)| \leq \varepsilon$. $\square$

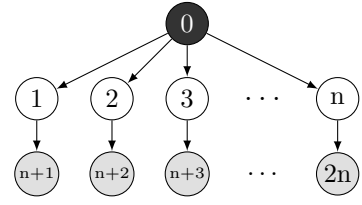

Figure 4: Credal Tree Used To Prove Theorem 2

Before proving our hardness results, we state and discuss some facts about the *partition problem*, which will be used later. The partition problem is stated as follows: given positive integers $z_1, \ldots, z_n$, decide whether there is $S \subseteq N := \{1, \ldots, n\}$ such that $\sum_{i \in S} z_i = \sum_{i \notin S} z_i$, where the notation $i \notin S$ denotes that $i \in N \setminus S$. This is a well-known NP-hard problem (Garey and Johnson, 1979). We define $v_i := z_i / z$, $i = 1, \ldots, n$, where $z := \sum_i z_i / 2$, and work w.l.o.g. with the partition problem using $v_i$ instead of $z_i$. Let $v_S := \sum_{i \in S} v_i$. Then, it follows for any $S$ that $v_S = 2 - \sum_{i \notin S} v_i$. Also, if an instance of the partition problem is a *yes-instance*, there is $S$ for which $v_S = 1$, whereas if it is a *no-instance*, then for any $S$, $|v_S - 1| \geq 1/(2z)$. Consider the function

$$h(v_S) = \frac{2^{-(v_S - 1)} + 2^{v_S - 1}}{2} \,. \tag{9}$$

Seen as a function of a continuous variable $v_S \in [0, 2]$, the function above is strictly convex, symmetric around one, and achieves the minimum value of one at $v_S = 1$. Thus, if the partition problem is a yes-instance, then $\min_S h(v_S) = 1$, while if it is a no-instance, then $\min_S h(v_S) \geq 2^{-1/(2z)-1} + 2^{1/(2z)-1} \geq 2^{(2z)^{-4}} > 1 + (2z)^{-4}/2 = 1 + 1/(32z^4)$, where the second inequality is due to Lemma 24 in (Mauá et al., 2012), and the strict inequality follows from the first-order Taylor expansion of $2^{(2z)^{-4}}$.

The following result shows that inferences under strong independence are hard even in credal trees.

**Theorem 2.** *Computing the GBR in credal trees under strong extension is NP-hard, even if all numerical parameters are rational numbers, and all variables are at most ternary.*

*Proof.* We use a reduction from the partition problem as previously described. We build a credal tree over variables $X_0, \ldots, X_{2n}$ with graph as in Figure 4. The root node is associated to the ternary variable $X_0$, with $\mathcal{X}_0 := \{1, 2, 3\}$ and uniform pmf $q(x_0) = 1/3$. The remaining variables are all Boolean. For $i = 1, \ldots, n$, specify the local sets $Q(X_i | x_0)$ as single-

tons $\{q(X_i|x_0)\}$ such that

$$q(x_i=1|x_0) = \begin{cases} 2^{-v_i}/(1+2^{-v_i}), & \text{if } x_0 = 1, \\ 1/(1+2^{-v_i}), & \text{if } x_0 = 2, \\ 1/2, & \text{if } x_0 = 3. \end{cases}$$

Finally, for $i = 1+n, \ldots, 2n$ specify the local credal sets such that $q(X_i|x_{i-n}) \in Q(X_i|x_{i-n})$ satisfies $q(x_i = 1|x_{i-n}) \in [\epsilon, 1]$ for all $x_{i-n}$, where $\epsilon = 2^{-n-3}/(64z^4)$. Consider the computation of the GBR with observed nodes $O = \{1+n, \ldots, 2n\}$, observation $\tilde{x}_O = (1, \ldots, 1) \in \mathcal{X}_O$, query node $q = 0$, and query $f(x_0) = -I(x_0=3)$. Using the results from (Antonucci and Zaffalon, 2006) about the *conservative inference rule*, one can show that $f$ is minimized at an extrema $p(X)$ such that $p(x_i=1|x_{i-n}=1) \neq p(x_i=1|x_{i-n}=0)$, that is, if $p(x_i=1|x_{i-n}=1)$ is chosen to be equal to $\epsilon$, then $p(x_i=1|x_{i-n}=0) = 1$, and vice-versa. Hence,

$$\mu = \min_p \mathbb{E}_p[f] = -\max_p p(x_0=3|\tilde{x}_O)$$
$$= -\max_p \left(\frac{1+\epsilon}{2}\right)^n \frac{1/3}{p(\tilde{x}_O)}$$
$$= -\max_S \frac{1}{g(a_S)},$$

where $g(a) = 1 + (1+a)\left(\frac{2}{1+\epsilon}\right)^n \prod_{i=1}^n 1/(1+2^{-v_i})$ is defined for any real number $a$, and $a_S := b_S + b_{N\setminus S} - 1$ and $b_S := \prod_{i\in S}(2^{-v_i}+\epsilon)\prod_{i\notin S}(1+2^{-v_i}\epsilon)$ are defined for all $S \subseteq N$. Note that $g(a_S) > 1 + (1+a_S)2^{-n}$. It follows from the Binomial Theorem that

$$2^{-v_S} \leq b_S \leq (2^{-v_S} + 2^n\epsilon)(1+\epsilon)^n$$
$$\leq (2^{-v_S} + 2^n\epsilon)(1+2n\epsilon)$$
$$\leq 2^{-v_S} + 2^{n+2}\epsilon$$

where we use the inequality $(1+r/k)^k \leq 1+2r$ valid for $r \in [0,1]$ and positive integer $k$ (Mauá et al., 2011, Lemma 37). Thus,

$$h(v_S) - 1 \leq a_S \leq h(v_S) + 2^{n+3}\epsilon - 1.$$

Now if the partition problem is a yes-instance, then $a_S \leq 1/(64z^4)$, while if it is a no-instance, we have that $a_S > 1/(32z^4)$. Hence, there is a gap of at least $1/(64z^4)$ in the value of $a_S$ between yes- and no-instances, and we can decide the partition problem by verifying whether $\mu \leq -1/g(\alpha)$, where $\alpha := 3/(128z^4)$. This proof shall be completed with the guarantee that we can approximate in polynomial time the irrational numbers used to specify the credal tree and $g(a)$ well enough so that $-1/g(\alpha)$ falls in the gap between the values of $\mu$ for yes- and no-instances. First, note that

$$g\left(\frac{1}{32z^4}\right) - g\left(\frac{1}{64z^4}\right) = \frac{1}{64z^4}\left(\frac{2}{1+\epsilon}\right)^n \prod_{i=1}^n \frac{1}{1+2^{-v_i}},$$

which is greater than $2^{-n}/(64z^4)$. The gap in the value of $\mu$ is at least

$$\frac{1}{g(1/(64z^4))} - \frac{1}{g(1/(32z^4))} = \frac{g(\frac{1}{32z^4}) - g(\frac{1}{64z^4})}{g(\frac{1}{64z^4})g(\frac{1}{32z^4})}$$
$$> \frac{g(\frac{1}{32z^4}) - g(\frac{1}{64z^4})}{g(\frac{1}{32z^4})^2}$$
$$> \frac{2^{-n}/(64z^4)}{(1+(1+\frac{1}{32z^4})2^{-n})^2}$$
$$> \frac{2^{-n}}{4 \cdot 64z^4}.$$

So we apply Lemma 1 with $\varepsilon = \frac{1}{2}\frac{2^{-n}}{4 \cdot 64z^4}$ and use the same rational numbers $q(x_i=1|x_o=2)$ as in the specification of the new network instead of the irrational values $1/(1+2^{-v_i})$ to approximate $g(\alpha)$, which guarantees that the gap will continue to exist. Alternatively, we can use the same argument as in Theorem 3 of (de Campos, 2011) to constructively find a suitable encoding for the numerical parameters and $g(\alpha)$. □

### 3.3 POLYTREES AND BEYOND

In general networks, it is still unclear which type of inferences depend on the irrelevance concept used. There is however one situation where we can show they coincide, and this is particularly important for the hardness results that we prove later on.

**Lemma 2.** *Consider a credal network of arbitrary topology, where all nodes are associated to precise pmfs apart from the root nodes, which are associated to vacuous credal sets. Then the result of the GBR for an arbitrary function $f$ of a variable $X_q$ associated to a non-root node $q$ and no evidence is the same whether we assume epistemic irrelevance or strong independence.*

*Proof.* Let $X_R$ be the variables associated root nodes (hence to vacuous local credal sets), and $X_I$ denote the remaining variables (which are associated to singleton local credal sets). The result of the GBR under epistemic irrelevance is given by

$$\mu = \min_{p(X)} \mathbb{E}_p[f] = \min_{p(X)} \sum_x p(x_I|x_R)p(x_R)f(x_q)$$
$$= \min_{p(X)} \sum_{x_R} p(x_R) \sum_{x_I} p(x_I|x_R)f(x_q)$$
$$= \min_{p(X_R)} \sum_{x_R} p(x_R)g(x_R),$$

where $g(x_R) := \sum_{x_I} \prod_{i \in I} q(x_i|x_{\text{pa}(i)})f(x_q)$, and $q(x_i|x_{\text{pa}(i)})$ are the single pmfs in the local credal sets of the non-root variables $X_I$. According to the last equality, $\mu$ is a convex combination of $g(x_R)$. Hence,

$$\mu \geq \min_{x_R} g(x_R) = \min_{x_R} \sum_{x_I} \prod_{i \in I} q(x_i|x_{\text{pa}(i)})f(x_q).$$

The rightmost minimization is exactly the value of the GBR under strong independence, and since the strong extension is contained in the epistemic extension, the inequality above is tight. □

The above result will be used in combination with hardness results for strong credal networks to demonstrate that computing the GBR under epistemic irrelevance is also hard. First, we focus on singly connected networks.

For polytrees, the next theorem shows that inferences in credal networks, either under epistemic irrelevance or strong independence, are NP-hard, even if variables are at most ternary. If we allowed variables to have any arbitrary finite number of states, then this result would follow from the proof of NP-hardness of inferences in polytrees given by de Campos and Cozman (2005), because the polytree presented there is similar to the one in Figure 5 with no evidence and query in the last (topological) node. By Lemma 2, we could use an inference in the epistemic polytree to solve the same inference, demonstrating that such inference is NP-hard too. In this paper we devise a stronger result, as the hardness is shown even when variables are at most ternary. For this purpose, we perform a polynomial-time reduction from the partition problem. A much similar reduction has been used to show that selecting optimal strategies in limited memory influence diagrams is NP-hard (Mauá et al., 2012). While the reduction used here closely resembles the reduction used in that work, due to a technicality we cannot directly use that result (mainly because the influence diagram could have multiple utility nodes, which would much complicate the reduction). Instead, we directly reduce an instance of the partition problem to a computation of the GBR without evidence in a credal polytree whose non-root nodes are all associated to precise pmfs and root nodes are associated to vacuous credal sets. By using this reduction in conjunction with Lemma 2, we prove the hardness of GBR computations also under epistemic irrelevance.

**Theorem 3.** *Given a credal polytree, computing the GBR with a function $f$ of the query variable $X_q$ and no evidence is NP-hard, whether we assume epistemic irrelevance or strong independence, even if all variables are (at most) ternary and all numbers are rational.*

*Proof.* We build a credal polytree with underlying graph as in Figure 5. The variables (associated to nodes) on the upper row are Boolean and vacuous, namely $X_1, \ldots, X_n$, while the remaining variables $X_{n+1}, \ldots, X_{2n+1}$ are ternary and associated to singleton local credal sets such that $Q(X_{n+1})$ contains a uniform pmf $q(x_{n+1}) = 1/3$, and, for $i = n+2, \ldots, 2n+1$, $Q(X_i|x_{i-1}, x_{i-n-1})$ contains the pmf $q(X_i|x_{i-1}, x_{i-n-1})$ as specified in Table 2. Consider a joint pmf $p(X)$ which is an extreme of the strong extension of the network. One can show that $p(x) = q(x_{n+1}) \prod_{i=n+2}^{2n+1} q(x_i|x_{i-1}, x_{i-n-1}) \prod_{i \in S} I(x_i = 1) \prod_{i \notin S} I(x_i = 0)$ for all $x$, where $S \subseteq N$. Thus, $p(x)$ is equal to $\frac{1}{3} \prod_{i=n+2}^{2n+1} q(x_i|x_{i-1}, x_{i-n-1})$ if $x_S = 1$ and $x_{N \setminus S} = 0$, and otherwise vanishes. It follows that $p(x_{2n+1} = 1) = \sum_x p(x) = (2^{-v_S})/3$ and $p(x_{2n+1} = 2) = (2^{v_S - 2})/3$. Let $\alpha := (1 + z^{-4}/64)/3$. By computing the GBR with query $f(x_{2n+1}) = I(x_{2n+1} = 1) + I(x_{2n+1} = 2)$ and no evidence, we can decide the partition problem, as $\min_p \mathbb{E}[f] = \min_S h(v_S)/3 \leq \alpha$, if and only if the partition problem is a yes-instance. According to Lemma 2, this result does not change if we assume epistemic irrelevance. It remains to show that we can polynomially encode the numbers $2^{-v_i}$. This is done by applying Lemma 1 with a small enough $\varepsilon$ computable in time polynomial in the size of the partition problem: $\varepsilon = 1/(3 \cdot 64z^4)$ suffices. □

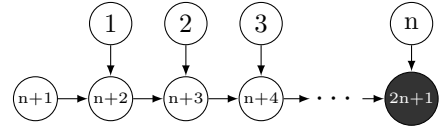

Figure 5: Polytree Used To Prove Theorem 3

Table 2: Local Pmfs Used To Prove Theorem 3

| $q(x_i\|x_{i-1}, x_{i-n-1})$ | $x_i = 1$ | $x_i = 2$ | $x_i = 3$ |
|---|---|---|---|
| $x_{i-1}=1, x_{i-n-1}=1$ | $2^{-v_i}$ | 0 | $1 - 2^{-v_i}$ |
| $x_{i-1}=2, x_{i-n-1}=1$ | 0 | 1 | 0 |
| $x_{i-1}=3, x_{i-n-1}=1$ | 0 | 0 | 1 |
| $x_{i-1}=1, x_{i-n-1}=0$ | 1 | 0 | 0 |
| $x_{i-1}=2, x_{i-n-1}=0$ | 0 | $2^{-v_i}$ | $1 - 2^{-v_i}$ |
| $x_{i-1}=3, x_{i-n-1}=0$ | 0 | 0 | 1 |

The hardness of inference in multiply connected credal networks under epistemic irrelevance comes from the fact that general credal networks under strong independence can always be efficiently mapped to credal networks under strong independence with all non-root nodes precise and vacuous root nodes. Because such inferences in general credal networks under strong independence are NP$^{\text{PP}}$-hard (Cozman et al., 2004), and because Lemma 2 demonstrates that marginal inferences without evidence in these networks are equivalent to the same inference in credal networks under epistemic extension, the hardness result is obtained also for epistemic networks. For completeness, we write the complete proof of such result using a reduction from the E-MAJSAT problem, since previous work has only provided a sketch of such proof (Cozman et al., 2004). The proof differs only slightly from the proof of NP$^{\text{PP}}$-hardness of MAP inference in Bayesian networks given by Park and Darwiche (2004).

**Theorem 4.** *Computing the GBR in credal networks under either epistemic irrelevance or strong independence is NP$^{\text{PP}}$-hard even if all variables are binary.*

*Proof.* The hardness result follows from a reduction from E-MAJSAT. Given a propositional formula $\phi$ over Boolean variables $Z_1, \ldots, Z_n$ and an integer $1 \leq k < n$, the E-MAJSAT is the problem of deciding whether there exists an assignment to $Z_1, \ldots, Z_k$ such that the majority of the assignments to the remaining variables $Z_{k+1}, \ldots, Z_n$ satisfy $\phi$. The reduction proceed as follows. Create a credal network over Boolean variables $X = (X_1, \ldots, X_k, X_{k+1}, \ldots, X_n)$ such that $X_1, \ldots, X_k$ are associated to root nodes and vacuous credal sets, and $X_{k+1}, \ldots, X_n$ are associated to non-root nodes and have uniform pmfs. The root variables act as selectors for the Boolean variables in the propositional formula. Each variable $X_i$ is associated to the Boolean variable $Z_i$ of the original formula $\phi$. Build one new binary variable $X_i$ (using a suitable sequence of numbers $i = n+1, n+2, \ldots$) for each operator in the Boolean formula $\phi$ of the E-MAJSAT problem such that $X_i$ has as parents its operands, that is, for logical operations $(X_a \wedge X_b)$ and $(X_a \vee X_b)$, with $a, b < i$, $X_i$ has as parents the two operands $X_a$ and $X_b$ and is associated to a singleton credal set containing the pmf $q(x_i|x_a, x_b) = I(x_i = x_a \circ x_b)$, where $\circ$ denotes the respective binary operation; for the operation $(\neg x_a)$, with $a < i$, $X_i$ has a single parent $X_a$ and is associated to a singleton local credal set containing the pmf $q(x_i|x_a) = I(x_i \neq x_a)$. There is more than one way to build such a network, depending on the order one evaluates the operations in the Boolean formula, and any valid evaluation order can be used. The final network encodes a circuit for evaluating the formula $\phi$.

Let $t$ be the last node in the network in topological order; variable $X_t$ represents the satisfiability of the whole formula. Consider a joint pmf $p(x) = \prod_{i \in N} q(x_i|x_{\text{pa}(i)})$, for some choice of pmfs from the extrema of local credal sets. By design, there is a single $x_{1:k} \in \mathcal{X}_{1:k}$ such that $\prod_{i=1}^{k} q(x_i)$ evaluates to one. Let $\tilde{x}_{1:k}$ be such (joint) value. We have that

$$p(X_t = 1) = \sum_{x \sim x_t} p(x_t = 1|x_I) p(x_I|x_R) \prod_{i=1}^{n} q(x_i)$$
$$= \frac{1}{2^{n-k}} \sum_{x \sim \tilde{x}_{1:k}} p(x_t = 1|x_I) p(x_I|x_R),$$

where $X_I$ are the non-root variables that represent the logical operations in the formula $\phi$ apart from $X_t$, $X_R = (X_1, \ldots, X_n)$ are the root variables associated to Boolean variables in $\phi$. The value of $p(x_t = 1|x_I) p(x_I|x_R)$ is equal to one if and only if the assignment $x_R$ satisfies $\phi$ (by construction) and is zero otherwise. Thus, $p(X_t = 1) = \#\text{SAT}/2^{n-k}$, where #SAT is the number of assignments to the variables $Z_{k+1}, \ldots, Z_n$ that satisfy $\phi$ with the first $k$ Boolean variables set to $\tilde{x}_{1:k}$. By maximizing over the possible choices of degenerate pmfs $q(X_i)$, $i = 1, \ldots, k$, we have that $\min_p p(x_t = 0) < 1/2$ if and only if there is an assignment to the first $k$ variables $Z_1, \ldots, Z_k$ such that more than half of the assignments to the remaining $n - k$ variables $Z_{k+1}, \ldots, Z_n$ satisfy the formula $\phi$. $\square$

## 4 CONCLUSION

In this paper, we have showed new computational complexity results for inference in credal networks under both epistemic irrelevance and strong independence. There are three main contributions. First, by exploiting the relations between these two irrelevance concepts in HMM-like credal networks, we have shown that predictive inferences under strong independence can be computed in polynomial time using the same algorithm developed for epistemic credal trees. To complement such result, we have proved that inferences with strong independence in general trees are NP-hard even if all variables are (at most) ternary, which shows that it is unlikely that more general polynomial-time algorithms for inferences under strong independence will ever exist. Moreover, using the relation between strong and epistemic irrelevance concepts in networks where the imprecision appears only in root nodes (defined by vacuous credal sets), we were able to prove that inferences in polytrees under epistemic irrelevance are NP-hard, even if all variables are (at most) ternary. This result closes the gap between known polynomial-time algorithms (which were known for trees and some polytrees) and potentially any more complicated network. To the best of our knowledge, these complexity results were all open, specially for the case of epistemic irrelevance.

## Acknowledgements

This work has been partly supported by the Swiss NSF grants no. 200020_137680/1 and 200020_134759/1, and the Hasler Foundation grant no. 10030.